\def\year{2021}\relax
\documentclass[letterpaper]{article} 
\usepackage{aaai21}  
\usepackage{times}  
\usepackage{helvet} 
\usepackage{courier}  
\usepackage[hyphens]{url} 
\usepackage{graphicx} 
\usepackage{natbib} 
\usepackage{caption}

\usepackage{hyperref}       
\usepackage{url}           
\usepackage{booktabs}      
\usepackage{amsfonts}     
\usepackage{nicefrac}  
\usepackage{microtype}      
\usepackage{amsmath}
\usepackage[linesnumbered, ruled]{algorithm2e}
\usepackage{multirow}
\usepackage{amsthm}
\usepackage{booktabs}
\usepackage{amssymb}
\usepackage{subfigure}
\usepackage{color}

\title{Learning to Augment for Data-Scarce Domain BERT Knowledge Distillation}

\author {
Lingyun Feng\textsuperscript{\rm 1,$ ^*$},
Minghui Qiu\textsuperscript{\rm 2,\thanks{Equal contributions.}},
Yaliang Li\textsuperscript{\rm 2},
Hai-Tao Zheng\textsuperscript{\rm 1,$ ^\dagger$},
Ying Shen\textsuperscript{\rm 3,\thanks{Corresponding authors.}}\\
}
\affiliations {
    \textsuperscript{\rm 1} Tsinghua University \textsuperscript{\rm 2} Alibaba Group
    \textsuperscript{\rm 3} Sun-Yat Sen University \\
fly19@mails.tsinghua.edu.cn,\\
\{minghui.qmh,yaliang.li\}@alibaba-inc.com,\\
zheng.haitao@sz.tsinghua.edu.cn,
sheny76@mail.sysu.edu.cn
}
\begin{document}

\maketitle

\begin{abstract}
Despite pre-trained language models such as BERT have achieved appealing performance in a wide range of natural language processing tasks, they are computationally expensive to be deployed in real-time applications. 
A typical method is to adopt knowledge distillation to compress these large pre-trained models (teacher models) to small student models.
However, for a target domain with scarce training data, the teacher can hardly pass useful knowledge to the student, which yields performance degradation for the student models. 
To tackle this problem, we propose a method to learn to augment for data-scarce domain BERT knowledge distillation, by learning a cross-domain manipulation scheme that automatically augments the target with the help of resource-rich source domains.
Specifically, the proposed method generates samples acquired from a stationary distribution near the target data and adopts a reinforced selector to automatically refine the augmentation strategy according to the performance of the student.
Extensive experiments demonstrate that the proposed method significantly outperforms state-of-the-art baselines on four different tasks, and for the data-scarce domains, the compressed student models even perform better than the original large teacher model, with much fewer parameters (only ${\sim}13.3\%$) when only a few labeled examples available.
\end{abstract}

\section{Introduction}
Pre-trained language models such as BERT~\cite{devlin2018bert}, XLNet~\cite{yang2019xlnet}, and RoBERTa~\cite{liu2019roberta} have demonstrated their extraordinary performance via fine-tuning on down-streaming natural language processing tasks~\cite{wang2018glue,williams2017broad,xu2018multi}.
However, the large number of parameters in these models leads to high storage and computational costs, which makes them a burden to be deployed in resource-constrained application scenarios such as real-time inference on mobile or edge devices.
A typical solution is to adopt knowledge distillation (KD)~\cite{hinton2015distilling} to reduce their storage and computation cost, and accelerate the inference time~\cite{sanh2019distilbert,sun2019patient,jiao2019tinybert,sunmobilebert}.  
The basic idea of knowledge distillation is to compress the large BERT model to a small student model while preserving the knowledge of teacher model. 
However, for a target domain with scarce training data, the teacher can hardly pass useful knowledge to the student, which yields performance degradation for the student models.

Data augmentation (DA) is a common strategy to deal with the data scarcity problem, which augments the target data by generating new data based on the labeled training sets. 
Nevertheless, designing an effective DA method for BERT knowledge distillation has been less explored. Existing augmentation methods for distillation are usually manually designed, such as thesaurus based synonym replacement~\cite{wang2015s,zhang2015character}, 
words replacement with paradigmatic relations~\cite{kobayashi2018contextual} or predictions of large language model~\cite{jiao2019tinybert,wu2019conditional}. 
Pre-defining such augmentation rules is time-consuming and hardly can find an optimal way to help knowledge distillation. 
It remains to be a challenging task to design an effective strategy to automatically augment useful data for the data-scarce domain.

In the light of this challenge, we propose a method to learn to augment data (L2A) for data-scarce domain BERT knowledge distillation that automates the process of data augmentation. 
Unlike prevailing data augmentation methods that pre-define heuristic rules, we automatically augment data and dynamically refine the augmentation strategy based on the feedback from the student model. 
The proposed method also leverages information from resource-rich domain data to help augment target data. Specifically, we adopt a reinforced selector to automate cross-domain data manipulation. 
The reinforced selector is a reinforcement learning policy network that controls the generation of cross-domain data based on the feedback from the student model. 
The student learns to mimic the behavior of the teacher with respect to teacher's dark knowledge (i.e., prediction logits) and intermediate hints
based on the augmented data.
Inspired by the reward augmented maximum likelihood~\cite{norouzi2016reward,ke-etal-2019-araml}, our generator is updated on the samples acquired from a stationary distribution weighted by the reinforced selector. Such stationary distribution is designed to guarantee that training samples are surrounding the real data, thus the exploration space of our generator is indeed restricted, leading to stable training. 

Experiments show that our model significantly outperforms the state-of-the-art baselines on different NLP tasks. 
The compressed student models even outperform the original large teacher models with much fewer parameters (only ${\sim}13.3\%$) when only a few labeled examples available.
This may due to the reason that, with the proposed L2A method, the student model removes the redundancy and noisy knowledge from the teacher model and only keeps the useful knowledge for the specific task. It echos the findings in recent studies \cite{tenney2019bertRedis,jawahar2019does} that BERT learns various knowledge from the large-scale corpus, while not all of them are useful for a specific downstream task.

In a nutshell, our main contributions are three-fold. 
\begin{itemize}
    \item  [1)] Instead of manually designing data augmentation methods for in-domain data,  our model can
leverage cross-domain information and automatically augment data according to the performance of the student model.
\item [2)] The proposed model generates samples from a stationary distribution with constrained exploration, which significantly reduces the search space and makes the training process stable. 
\item [3)] Experiments on different tasks show our model consistently outperforms the baselines, including the state-of-the-art distillation methods, DA methods, and even the teacher model.
\end{itemize}

\section{Related Work}
\textbf{Knowledge distillation} has proven a promising way to compress large models while maintaining accuracy. It transfers knowledge from a large model or an ensemble of neural networks (i.e., teacher) to a single lightweight model (i.e., student). The study in~\cite{hinton2015distilling} first proposes to use the soft target distributions of the teacher model to train the student model. Intermediate representations from hidden layers of the teacher are also useful for the student~\cite{Romero15-iclr}.
DistillBERT~\cite{sanh2019distilbert} uses the soft label and embedding outputs of the teacher to train the student. 
Recent work TinyBERT~\cite{jiao2019tinybert} and MobileBERT~\cite{sunmobilebert} further consider self-attention distributions and hidden states to train the student. $\text{BiLSTM}_\text{SOFT}$~\cite{tang2019distilling} distills fine-tuned BERT into a LSTM model. 
However, for a domain with scarce training data, the teacher can hardly pass useful knowledge to the student, which yields performance degradation for the student models.
Both TinyBERT~\cite{jiao2019tinybert} and $\text{BiLSTM}_\text{SOFT}$~~\cite{tang2019distilling} use data augmentation to improve the distillation performance. But these handcraft augmentation methods are time-consuming and may not perform well during training.

\textbf{Data augmentation} is a ubiquitous technique to augment the target data by generating new data from existing training data, with the objective of improving the performance of the downstream tasks. Most of the studies are based on heuristics such as synonym replacement, random insertion, random swap, and random deletion~\cite{zhang2015character,wei2019eda}. Generation based approaches are also studied, including variational Auto-Encoder (VAE)~\cite{kingma2013auto}, round-trip translation~\cite{yu2018qanet}, paraphrasing~\cite{kumar2019submodular} and data noising~\cite{xie2017data}, which try to generate sentences from a continuous space with desired attributes of sentiment and tense. Recently, pre-trained language models are adopted to synthesize new labeled data.
However, the exploration space is usually huge and the quality of the sentences generated by these methods may not be satisfactory.
Contextual augmentation is introduced in~\cite{kobayashi2018contextual,wu2019conditional}, where bi-directional language models or fine-tuned BERT~\cite{devlin2018bert} are used to replace words in the original sentence with other words based on context information.
However, these methods are designed manually and cannot dynamically adjust the augmentation strategy during the training stage, which could result in suboptimal performance. 

Automated data augmentation has been proposed in \cite{ratner2017learning,cubuk2019autoaugment}, however, they are designed for image augmentation, which are not suitable for text augmentation due to the complexity nature of language. Recently, adaptive text augmentation have been studied in~\cite{hu2019learning}. However, it is not suitable for BERT compression and can not handle cross-domain data manipulation. Meanwhile, its data manipulation is designed for text classification which is not suitable for text-pair tasks such as PI and NLI studied in this paper. Besides, existing methods on data augmentation only focus on a particular target domain data. In this study, we treat data augmentation as a learning task to ``generate'' new data to help the target domain leveraging cross-domain information.

\section{Learning to Augment}
\begin{figure*}[t!]
  \centering
  \includegraphics[width=0.9\linewidth]{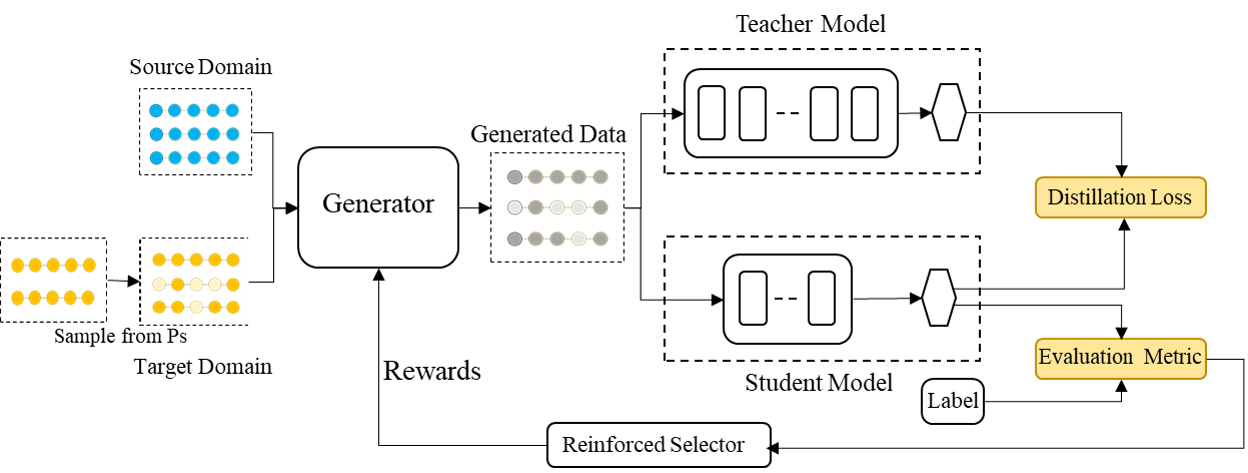}
  \caption{Overview of the proposed Learning to Augment (L2A) method. The generator generates augmented data based on both source and target domain data from a statistic stationary distribution~($P_s$). The reinforced selector selects useful augmented data to help the task of knowledge distillation and updates its policy according to the student network performance. }
\label{fig:overview}
\end{figure*}

We present an overview of the proposed Learning to Augment (L2A) method in Figure~\ref{fig:overview}. Unlike prevailing data augmentation methods that pre-define heuristic rules for a particular target domain, we automatically augment data from both a source domain and a target domain, and dynamically refine the augmentation strategy based on feedback from the student module.

Specifically, our learning framework mainly consists of four components, i.e., a teacher module, a student module, a data generator, and a reinforced selector.
The data generator generates data from both source and target domains for the teacher module to guide the student module. 
The reinforced selector refines the augmentation strategy of the data generator based on the performance of the student module.
Those modules work interactively to jointly improve the performance of the student module for a target domain.

\subsection{Knowledge Distillation}
\label{sec:kd}
Before we dive into our method, we first introduce the process of knowledge distillation. The distillation process aims to transfer the knowledge of a large teacher network to a small student network. 
The objective is defined as follows:
\begin{align}
     \mathcal{L}_{KD} &=\sum_{x\in \mathcal{X}}L(f^s(x),f^t(x)),
\label{eq:KD loss}
\end{align}

where $f^s$ and $f^t$ represent the features of student and teacher models respectively. 
$L(\cdot)$ is a loss function that evaluates the difference between the teacher and student models. Inspired by the success of transformer networks such as BERT~\cite{sun2019patient,jiao2019tinybert,wang2020minilm}, our distillation model is based on the BERT network~\cite{vaswani2017attention}. We consider three types of distillation strategy: $\mathcal{L}_{att}$ based on attention information, $\mathcal{L}_{hidden}$ on intermediate hidden representations, and $\mathcal{L}_{dark}$ on the prediction outputs or dark knowledge, detailed as follows:

\begin{equation}
    \begin{aligned}
         \mathcal{L}_{att}&=\frac{1}{h}\sum_{i=1}^{h}\mathit{MSE}(A_i^s - A_i^t),\\
    \mathcal{L}_{hidden}&=\mathit{MSE}(H^s W - H^t),
    \end{aligned}
\end{equation}
where $A_i$ represents the attention matrix corresponding to the $i$-th self-attention head of the last BERT layer and $h$ is the number of attention heads. $H^s, H^t$ denotes the output of the last layer of student network and teacher network, respectively.
$W$ denotes a transformation matrix that 
transforms the hidden states of the student network into the same space as the teacher network’s states.

For dark knowledge based distillation, we penalize the soft cross-entropy loss between the student network’s logits against the teacher’s logits as follows:
\begin{equation}
     \mathcal{L}_{dark}=-\sum_i \frac{\exp(g_i^t/T_{KD})}{\sum_j \exp(g_j^t/T_{KD})} \log \frac{\exp(g_i^s/T_{KD})}{\sum_j \exp(g_j^s/T_{KD})},
\end{equation}
where $g^s$ and $g^t$ are the logits from the student and teacher respectively. $T_{KD}$ denotes the temperature value which controls the smoothness of the output distribution. Note that for the regression problem, the above loss is reformulated as the mean square error between the student's and the teacher's logits.

We combined the above three types of loss as our final KD loss, namely:
$\mathcal{L}_{KD} = \mathcal{L}_{att} + \mathcal{L}_{hidden} + \mathcal{L}_{dark}$.

\subsection{The Proposed Method}
\label{sec:prop}
For a knowledge distillation task, training in a data-scarce domain may not be sufficient for the teacher model to fully express its knowledge~\cite{ba2014deep}. 
Therefore, we propose to augment the training set to help effective knowledge distillation in data-scarce domains. Specifically, we seek to train a generative model $G_\theta(z|x)$ to generate augment samples for the student model to better learn from the teacher. Here, $z$ refers to the generated sample, and $x$ refers to the original data. Thus the distillation loss is reformulated as follows:
\begin{equation}
     \mathcal{L}_{KD}=\mathbb{E}_{z \sim G_\theta(z|x)} L(f^s(z), f^t(z)).
\label{eq:KD loss2}
\end{equation}

Suppose we have an explicit metric $R_\phi(z)$ to evaluate an augmented sample $z$. The generator $G_\theta$ can be optimized via reinforcement learning, formulated as follows:
\begin{equation}
     \mathcal{L}_{RL,\theta}=-\mathbb{E}_{z \sim G_\theta(z|x)} [R_\phi(z)] - \alpha \mathbb{H}(P_{G_\theta(z|x)}),
\label{eq:rl_loss}
\end{equation}
where the first term is to maximize the expected rewards and the second term is to regularize the policy to generate diverse samples. $\alpha$ is a temperature hyper-parameter to balance these two terms.

The task can be viewed as a sequence generation task, in which at each step the generator acts over a huge discrete action space~\footnote{Vocabulary size is usually large} in the language model that makes the exploration of policy quite inefficient.
In order to alleviate the exposure bias problem and model collapse in text generation, we define 
an exponential payoff distribution to connect RL loss with RAML loss inspired by Reward Augmented Maximum Likelihood (RAML)~\cite{norouzi2016reward,ke-etal-2019-araml}. The loss is rewritten as:

\begin{equation}
    \begin{aligned}
    \mathcal{L}_{RAML,\theta}
    &= KL(Q_\phi(z) )||P_{G_\theta}(z|x))+constant\\
    &\propto -\mathbb{E}_{z \sim Q_\phi(z)} [\log P_{G_\theta}(z|x)],
\label{eq:RL2}
\end{aligned}
\end{equation}
where the second expression omits constant terms.
Here the exponential payoff distribution $Q_\phi(z)$ captures the relative reward of the generated sample compared with all the possible
perturbed samples. Hence we define $Q_\phi(z) = \frac{1}{Z} \exp(R_\phi(z)/\alpha)$, where $\alpha$ is the same as in Eq.~\ref{eq:rl_loss}, which controls the degree of regularization. $Z$ denotes the accumulation over all possible samples. 

As \cite{norouzi2016reward} shows, the global minimum of $L_{RL,\theta}$, i.e. the optimal regularized expected reward, is achieved when the model distribution matches the exponentiated payoff distribution, i.e.$P_{G_{\theta}}(z|x)=Q_{\phi}(z)$. Thus we first sample proportionally to its exponentiated scaled reward. Since directly sampling from the exponential payoff distribution $Q_\phi(z)$ is intractable, we define a stationary distribution $P_s(z)$ and then introduce importance sampling to separate sampling process. We then refine the loss as follows. 
\begin{equation}
    \begin{aligned}
        \mathcal{L} &= -\mathbb{E}_{z \sim Q_\phi(z)} [\log P_{G_\theta}(z|x)] \\
&= -\int_{z}\log P_{G_\theta}(z|x) Q_{\phi}(z) dz \\
&= -\int_{z}\log P_{G_\theta}(z|x) \frac{Q_{\phi}(z)}{P_s(z)} P_s(z) dz \\
& = -\mathbb{E}_{z \sim P_s(z)} [W_\phi(z) \log P_{G_\theta} (z|x)],
    \end{aligned}
\end{equation}
where $W_\phi(z)=\frac{Q_{\phi}(z)}{P_s(z)} \propto R_\phi(z)$.  
To optimize this loss function, we first construct the fixed distribution $P_s(z)$ to get samples, then devise the proper reward function $R_\phi(z)$ to train the augmentation model in a stable and effective way.

\begin{equation}
    \begin{aligned}
\mathcal{L}_{L2A,\theta} &= -\mathbb{E}_{z \sim P_s(z)} [W_\phi(z) \log P_{G_\theta} (z|x)] \\
&\propto -\mathbb{E}_{z \sim P_s(z)} [R_\phi(z) \log P_{G_\theta} (z|x)] 
\end{aligned}
\end{equation}

Intuitively this reward function $R_{\phi}(z)$ should encourage the generator to generate samples with large sampling probability $P_s(z)$ and also be helpful for knowledge distillation. Hence $R_{\phi}(z)$ is defined by two terms: $R_{\phi}(z) =logP_s(z) +\pi_{\phi}(z)$, where the former address the sampling probability and the latter is controlled by the reinforced selector to guarantee the generated samples is helpful for knowledge distillation. To examine its helpfulness for distillation, we adopt a reinforced selector to learn a selection network $\pi_\varphi(z) \in [0, 1]$ for each augmented $z$. Thus, the learning task is reformulated as follows:

\begin{align}
    \label{eq:l2a}
\mathcal{L}_{L2A,\theta} 
&\propto -\mathbb{E}_{z \sim P_s(z)} [R_\phi(z) \log P_{G_\theta} (z|x)]  \\
&= -\mathbb{E}_{z \sim P_s(z)} [(\log P_s(z) + \pi_\varphi(z)) \log P_{G_\theta} (z|x)]. \nonumber
\end{align}

The search space $P_s(z)$ and $\pi_\varphi(z)$ in the reward function $R_{\phi}(z)$ are detailed in the following sections.

\subsubsection{Constrained Search Space}
\label{sec:space}
The stationary distribution $P_s(z)$
is designed to guarantee that training samples are surrounding the real data, thus the exploration space of our generator is indeed restricted, resulting in more stable training.
Based on this intuition, we sample from $P_s(z)$ where $P_s(z)=\mathbb{E}_{x}[P_s(z|x)]$ by stratified sampling for data augmentation, where we first select a particular distance $d$, then sample the position $p$ for substitution, and fill a word $w$ into the position. So $P_s(z|x)$ can be derived as:
 
\begin{equation}
\begin{aligned}
P_s(z|x) = P(d,o,w|x) =P(d|x)P(o|d,x)P(w|o,d,x).
\end{aligned}
\label{eq:sample}
\end{equation}

We first sample an edit distance $d$ by $P(d|x)$, then select position $o$ for substitution based on the sampled edit distance by $P(o|d,x)$. Then we determine the probability of word substitution by $P(w|o,d,x)$.
Let $|V|$ be the size of vocabulary and $c(e,m)$ denote the number of sentences at an edit distance $e$ from a sentence of length $m$, i.e., $c(e,m)= \tbinom{m}{e}\cdot (|V|-1)^e$. Follow \cite{norouzi2016reward}, we reweight the counts and normalize the result:
\begin{equation}
    P(d|x)=\frac{exp\{-d/\alpha\}c(d,m)}{\sum_{e=0}^mexp\{-e/\alpha\}c(e,m)},
    \label{eq:param-a}
\end{equation}
where $\alpha$ here is defined the same as the temperature hyperparameter in Eq.~\ref{eq:rl_loss}, which controls the degree of regularization and restricts the search space surrounding the original
sentence. 
A larger $\alpha$ brings more samples with long edit distances. We test the effect of $\alpha$ in the experiment section.

Then we randomly choose $d$ distinct positions in a sentence to be replaced by new words, the probability of choosing the position $o$ can be calculated as $P(o|d,x)=d/m$. 

Besides using $\alpha$ to restrict the search space surrounding the original sentence, we also adopt a constrained sampling strategy to constrain the exploration of new samples.
Inspired by the 
success of BERT~\cite{devlin2018bert} for NLP tasks, we leverage the rich contextual and semantic information within BERT to generate semantically coherent variants of the ground truths. Specifically, for $p(w|o, d, x)$, we adopt BERT based generator by masking a position $o$ with a special token $[mask]$ and using BERT to predict the corresponding word on the position to generate a new sentence. We used a softmax-temperature:
\begin{equation}
    P(w|o,d,x)=\frac{\exp(P_{\text{BERT}}(w)/T)}{\sum_j \exp(P_{\text{BERT}}(w_j)/T)},
    \label{eq:param-t}
\end{equation}
where $P_{\text{BERT}}(w)$ denotes the probability of generating word $w$ based on the BERT model. T controls the exploration degree, where a higher value for T produces a softer probability distribution over candidate words.
By controlling $\alpha$ and $T$, we restrict the exploration space of the generator to a set of grounded samples close to the ground truths. We discuss their effect in the experiment section.

Note that $P_s(z)$ can be drawn from both the source domain and target domain. Since the source domain is rather large, we set the sample size to 1 for source domain and 20 for target domain in the experiments, i.e. we use the full source data and generate 20 samples for each instance in target domain. The effect of different sample size are shown in the experiment section. Next, we use a data augmentation policy to sample from the candidates to generate the final augmented data.

\subsubsection{Reinforced Selector}\label{sec:ctrl}
We adopt the reinforced selector to provide an assessment for each augmented data. We seek to update the selection network automatically using the feedback from the student model results. Formally, we define the state of our model as the outputs of the teacher and student model and the action as a binary decision (0 or 1) on each input sample. 
Furthermore, to speed up training, we treat an epoch as an episode and each batch as a step to act on.
We define the reward at time step $t$ based on the evaluation metric which measures difference in performance before and after the student model updates:
\begin{equation}
    r_t=L(y_i,f^s(x_i))-L'(y_i,f^s(x_i)),
\label{eq:delayed}
\end{equation}
where $L(y_i,f^s(x_i))$ denotes the evaluation results of the updated model, and $L'(y_i,f^s(x_i))$ denotes previous the evaluation results. For classification tasks, $L$ is set to the accuracy of the target domain validation data. For regression tasks, $L$ is set to the correlation coefficient between the predicted score and the ground truth score.
In contrast to conventional reinforcement learning, our model is updated in batches in order to improve the model training efficiency. For each batch in an episode, the accumulated reward is defined as:
\begin{equation}
   r(\tau)=\sum_{k=0}^{T-t}\gamma^k r_{t+k},
\label{eq:total}
\end{equation}
where $\gamma$ is a discount factor. $\tau$ denotes the sequence of actions through time $T$. 

Our selector executing actions according to policy $\pi_{\varphi}$, which aims to maximize the expected discounted sum of rewards of over trajectory $\tau$:
\begin{equation}
\varphi^*=\mathop{\arg\max}_{\varphi} \mathbb{E}_{\tau \sim \pi_\varphi(\tau)}r(\tau).
\label{eq:policy}
\end{equation}

\begin{algorithm}[t!]
\caption{Learning to Augment for Data-Scarce Domain BERT Compression}\label{algorithm}
\SetKwInOut{Require}{Require}
 \Require{training set $\mathcal{D}$, validation data $\mathcal{D}^v$}
 
Initialize the KD module and reinforced selector;\\

Construct the distribution $P_s$ using Eq.~\ref{eq:sample};\\
Sample from $P_s$ and get training data $D'$;\\
\For{each batch $x_b$ in $D'$}{
  Obtain teacher model output $f^t(x_b)$, student model output $f^s(x_b)$ and get state $s_b$;\\
  Augment data using Eq.~\ref{eq:l2a} and obtain action $a_b$; \\
  Update the student model using Eq.~\ref{eq:KD loss};\\
  Obtain the reward $r_b$ using Eq.~\ref{eq:delayed};\\
  Store $(s_b,a_b,r_b)$ in episode history H;
}
\For{each turple $(s_b,a_b,r_b)$ in the history H}{
Obtain the accumulated reward using Eq.~\ref{eq:total}; \\
Update policy $\pi_{\varphi}$ using Eq.~\ref{eq:policy}.
}
\end{algorithm}

\subsection{Training}
In all, we present the training algorithm in Algorithm~\ref{algorithm}.
Unlike manually designed data augmentation methods for in-domain data, our model can learn cross-domain manipulations and automatically augment data according to the feedback of the student model. Besides, the proposed method generates samples from a stationary distribution with constrained exploration space which significantly reduces the search space and makes the training process stable.

\section{Experiments}
We conduct experiments on four NLP tasks
to examine the efficiency and effectiveness of the proposed method. 
\subsection{Datasets}
\noindent\textbf{Natural Language Inference (NLI).}  This is a task to examine whether a hypothesis can be inferred from a premise~\cite{bowman2015large}. We use MultiNLI~\cite{williams2017broad} as the source domain and SciTail~\cite{khot2018scitail} as the target. The former is a large crowd-sourced benchmark corpus from a wider range of text genres, while the latter is a recently released challenging textual entailment dataset collected from the science domain. 

\noindent\textbf{Paraphrase Identification (PI).} This is a task to examine the relationship, i.e., a paraphrase or not, between two input text sequences.
We treat the Quora question pairs~\footnote{\url{www.kaggle.com/c/quora-question-pairs}} as the source domain and a paraphrase dataset made available in CIKM AnalytiCup 2018~\footnote{\url{https://tianchi.aliyun.com/competition/introduction.htm?raceId=231661}} as the target. 
The former is a large scale dataset that covers a variety of topics, while the latter consists of question pairs in the E-commerce domain. We follow the study in~\cite{qu2019learning} for data preprocessing.

\noindent\textbf{Text classification}. We treat SST-2~\cite{socher2013recursive} as source domain and RT~\cite{pang2005seeing} as target. SST-2~\cite{socher2013recursive} consists of sentences extracted from movie reviews with human annotations of their sentiments. 
RT~\cite{pang2005seeing} is a movie review sentiment dataset contains a collection of short review excerpts.

\noindent\textbf{Review helpfulness prediction.} 
This task is to examine the helpfulness score of a given review. Due to the high volume of reviews in E-commerce sites, its an important task that draws increasing attention.
We use the Electronics domain in the Amazon review dataset~\cite{mcauley2013hidden} as source data and the Watches domain as the target. We follow the study~\cite{chen2018cross} for data preprocessing.

\begin{table*}[t!]
\caption{Evaluation results on NLI, PI, text classification, and regression tasks. ``-'' means the method is not suitable for this task. P. and S. denotes Pearson and Spearman correlation, respectively.}
\centering
\begin{tabular}{l r c c c c c c c c}
\toprule \multirow{2}{*}{Method} & Model
 & \multicolumn{2}{c}{NLI} & \multicolumn{2}{c}{PI} & \multicolumn{2}{c}{Text Classification} & \multicolumn{2}{c}{Regression Task} \\ \cline{3-10} 
& Size & ACC    & F1     & ACC    & F1     & ACC    & F1     & P. & S. \\ \midrule
Student-FT &14.5M& 0.7380  & 0.6928 & 0.8844 & 0.7435 & 0.7188 & 0.7309 & 0.3878  & 0.3225    \\
$\text{BiLSTM}_\text{SOFT}$        & 10.1M& 0.5890  & 0.5006 & 0.8622 & 0.7009 & 0.4839 & 0.6522 &    -     &   -  \\
DistilBERT    & 52.2M & 0.6891 & 0.5648 & 0.8991 & 0.7775 & 0.6776 & 0.6966 & 0.4048  & 0.3343    \\
$\text{BERT}_4$-PKD& 52.2M& 0.5809 & 0.5819 & 0.9041  & 0.7956 & 0.6173 & 0.5189 & 0.4466  & 0.3778    \\
$\text{BERT}_6$-PKD &67M & 0.6980 & 0.6201 & 0.9060  & 0.8040 & 0.6370 & 0.6311 & 0.4482  & 0.3923    \\
MINILM        &33M & 0.7512 & 0.6314 & 0.9024 & 0.7858 & 0.7020  & 0.7022 & 0.4441  & 0.4132    \\
TinyBERT        &14.5M &0.7319 & 0.6143 & 0.8787 & 0.7274 & 0.7235 & 0.7392 &  0.2653       & 0.2139          \\ \hline
EDA              &\multirow{2}{*}{14.5M} &0.7465      &   0.6375    &  0.9030      &    0.7920    &    0.7254    &  0.7428      &    0.4554     &   0.3887        \\
CBERT            &  &  0.7469  &    0.6820   &   0.8925     &     0.7654   &    0.7366    &     0.7020   &    0.4680     &    0.3891       \\ \hline
L2A   & 14.5M & \textbf{0.7827} & \textbf{0.7152}  &    \textbf{0.9195}   &  \textbf{0.8275}  & \textbf{0.7798} & \textbf{0.7614} &  \textbf{0.4852}    &  \textbf{0.4204}      \\ \bottomrule
\end{tabular}
\label{tab:results}
\end{table*}

To mimic data-scarce domains, we subsample a small training set from the target domain for NLI and text classification tasks by randomly picking 40 instances for each class, and take 1\% of the original data as our training data for review helpfulness prediction task. Since the target domain in the PI task is relatively small, we keep it unchanged.

\subsection{Baselines}
We compare the proposed method with several state-of-the-art BERT compression methods including $\text{BiLSTM}_\text{SOFT}$~\cite{tang2019distilling}, DistilBERT~\cite{sanh2019distilbert}, BERT-PKD~\cite{sun2019patient}, TinyBERT~\cite{jiao2019tinybert} and 
MINLILM~\cite{wang2020minilm}~\footnote{We use the uncased version from \url{https://github.com/microsoft/unilm/tree/master/minilm}. The number of layers=12, hidden size=384 and head number=12}. 
Note that TinyBERT and $\text{BiLSTM}_\text{SOFT}$ also conduct data augmentation for training. We also compare with two state-of-the-art DA methods, namely Easy Data Augmentation (EDA)~\cite{wei2019eda} and Conditional BERT (CBERT)~\cite{wu2019conditional} while using the same KD model as in our model to make fair comparisons. EDA is a simple but effective rule-based data augmentation framework for the text, which includes synonym replacement, random insertion, random swap, and random deletion. CBERT uses a language model that generates new variants semantically close to the original ones.
To keep comparisons fair, the number of generated augmented sentences per original
sentence is same for all comparing DA methods.

\subsection{Implementation Details}

we use the $\text{BERT}_\text{BASE}$ \cite{devlin2018bert} as the teacher model. For teacher model, the number of layers is 12, hidden size is set to 768, feed-forward/filter size is 3072 and head number is 12.
We initialize our student model with  $\text{BERT}_\text{Tiny}$~\footnote{\url{https://github.com/huawei-noah/Pretrained-Language-Model/tree/master/TinyBERT} (2nd version)}. For the student model, the number of layers is 4, hidden size is 312, feedforward/filter size is 1200 and head number is 12. 
All models are implemented with PyTorch~\cite{paszke2019pytorch} and Python 3.6.
We set the maximum sequence length to 128.
We tune the temperature $\alpha$ from \{0.6,0.7,0.8,0.9,1.0\} and choose $\alpha=0.6$ for the best performance. We tune $T$ from \{1,2,4,8\} and choose $T=1$ for the best performance. The sample size is 20 for target domain and 1 for source domain. 
The batch size is chosen from \{8,16,32\} and the learning rate is tuned from \{2e-5, 3e-5, 5e-5\}.
For the reinforced selector,
We use Adam optimizer~\cite{kingma2015adam} with the setting $\beta_1=0.9$, $\beta_2=0.998$.
The size of the hidden layer of the policy network is 128. The learning rate is set to 3e-5.
Note that, fine-tuned models are also compared, where the teacher and student models that are fine-tuned with the target domain data, denoted as Teacher-FT and Student-FT respectively.

\subsection{Evaluation Results}
As shown in Table~\ref{tab:results}, we have several observations. 

1) L2A significantly improves over the base student model by 6\% on average and consistently outperforms baselines in all tasks. This indicates that the proposed learning framework can effectively improve the performances of small student models for all these downstream tasks.

2) L2A outperforms the state-of-the-art BERT knowledge distillation methods, which indicates the L2A can effectively improve knowledge distillation performance by generating useful augmented instances. 

3) Our method also outperforms existing heuristic DA methods which shows that heuristic rules do not fit the task or datasets well. In contrast, learning-based augmentation has the advantage of automatically generating useful samples to improve model training.

\begin{table*}[t!]
\caption{Comparisons with the teacher and student fine-tuning on all datasets. 
P. and S. mean Pearson and Spearman correlation respectively. The bold and underlined numbers represent the best and 2nd best results respectively.}
\centering
\begin{tabular}{l r l l l l l l l l}
\toprule \multirow{2}{*}{Method} & Model
 & \multicolumn{2}{c}{NLI} & \multicolumn{2}{c}{PI} & \multicolumn{2}{c}{Text Classification} & \multicolumn{2}{c}{Regression Task} \\ \cline{3-10} 
& Size & ACC    & F1     & ACC    & F1     & ACC    & F1     & P. & S. \\ \midrule
Teacher-FT & 109M & 0.7639 & 0.6935 & \textbf{0.9225} & \textbf{0.8359} & 0.7573& \underline{0.7604} & \textbf{0.4874}  & \textbf{0.4264}    \\
Student-FT &14.5M& 0.7380  & 0.6928 & 0.8844 & 0.7435 & 0.7188 & 0.7309 & 0.3878  & 0.3225    \\ \hline
L2A$_{w/o \ tgt}$     & \multirow{3}{*}{14.5M}&     \underline{0.7714}&\underline{0.7025}&0.8549&0.6730&\underline{0.7610}&0.7442&0.4715&0.4119    \\
L2A$_{w/o \ src}$     &  & 0.7615 & 0.6955 & 0.9144 &  0.8165 & 0.7526 & 0.7523 &    0.4757   & 0.3992        \\
L2A   && \textbf{0.7827} & \textbf{0.7152}  &    \underline{0.9195}   &  \underline{0.8275} & \textbf{0.7798} & \textbf{0.7614} & \underline{0.4852} & \underline{0.4204} \\ \bottomrule
\end{tabular}
\label{tab:results2}
\end{table*}

\subsection{Comparison with the Teacher and Student Model}
We further compare different variants of our method with the teacher model in Table~\ref{tab:results2}. We can observe that:

1) L2A outperforms both L2A$_{w/o \ src}$ and L2A$_{w/o \ tgt}$ in all tasks, showing that data augmentation based on either source domain or target domain information can help to improve model performance.

2) Clearly, there is a large performance gap between the teacher model Teacher-FT and student model Student-FT due to the big reduction in model size. The fact that both L2A and L2A$_{w/o \ src}$ manage to bridge the gap between student and teacher model shows the proposed methods can effectively improve knowledge distillation performance by generating useful augmented instances. 

3) Compared with the teacher $\text{BERT}_\text{BASE}$, the compressed L2A model is 7.5x smaller (with only $\sim 13.3\%$ parameters), while maintaining comparable performance. L2A even outperforms the teacher model by 2\% on NLI and text classification tasks. This shows for data-scarce domain, proper data augmentation can significantly improve the student performance and even achieve comparable or better performance than the teacher.

Recall that BERT learns various knowledge from the large-scale corpus, while only certain parts of the learned knowledge are needed for a specific task~\cite{tenney2019bertRedis,jawahar2019does}. 
In the proposed L2A method, the student can help to only learn the needed knowledge from the large BERT teacher model and build a compact yet effective model to improve the downstream task performance.

\begin{figure}[t!]
  \centering
  \includegraphics[width=0.9\linewidth]{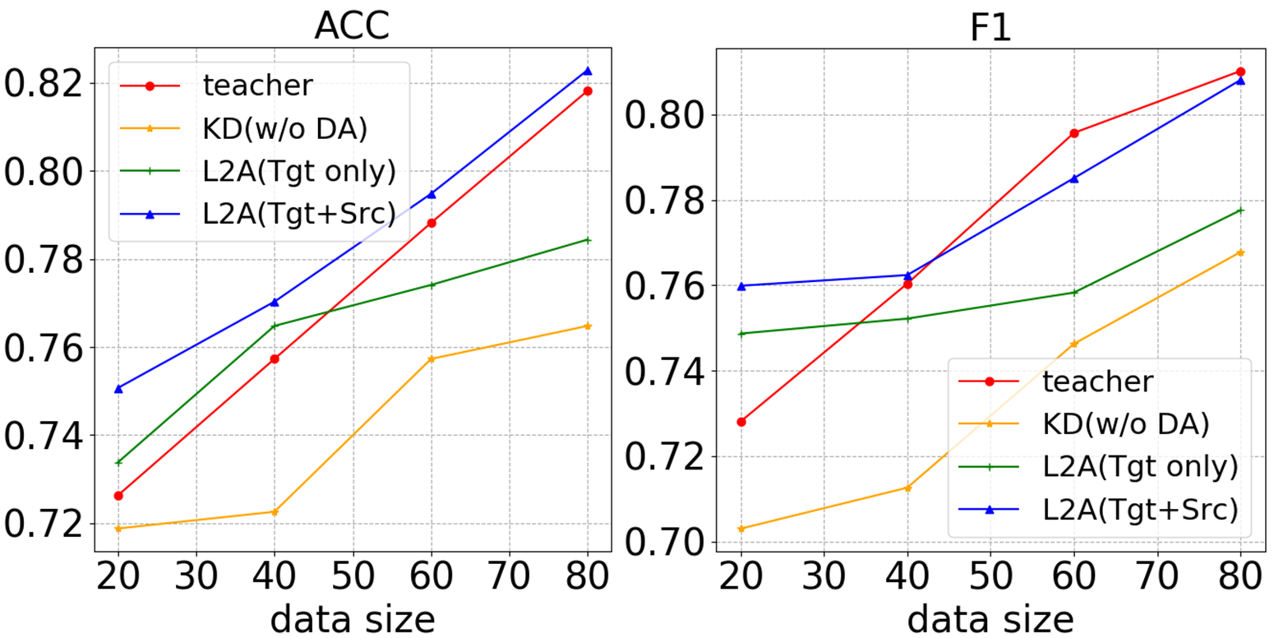}
  \caption{Ablation study on different target domain data sizes.}
  
\label{fig:data_size}
\end{figure}

\begin{figure}[t!]
  \centering
  \includegraphics[width=0.9\linewidth]{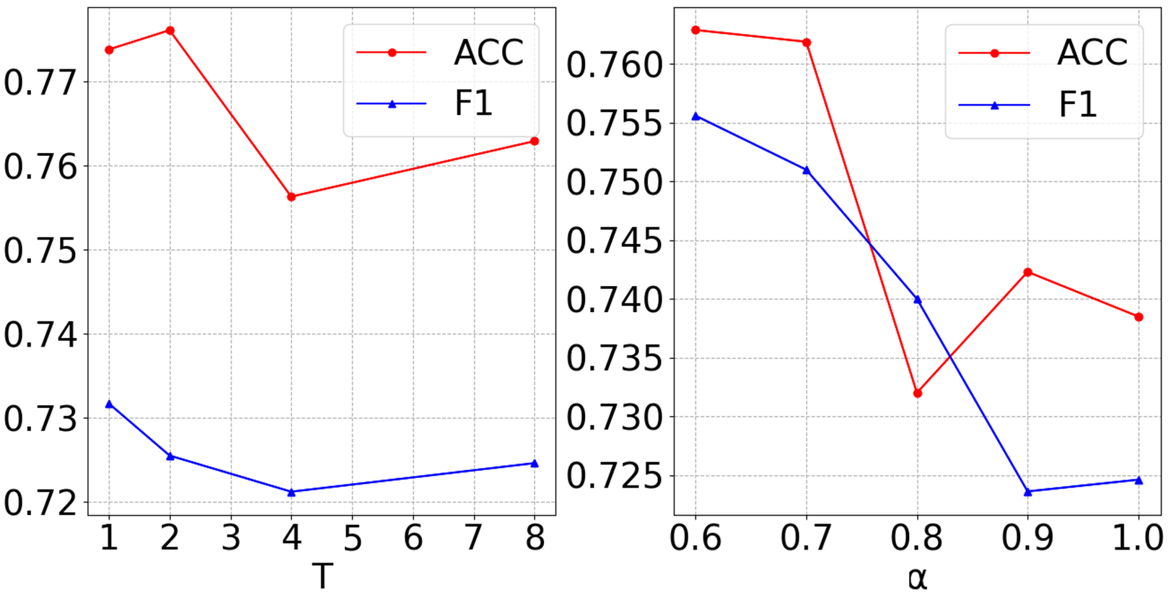}
  \caption{Ablation study on different  temperature values.}

\label{fig:temp}
\end{figure}

\subsection{Ablation Study}
\noindent\textbf{Effects of data size}. We compare over varied numbers of trained samples: $20, 40, 60, 80$ for each class to examine model performance w.r.t. different domain data size. As in Figure~\ref{fig:data_size}, L2A consistently improves the performance of the knowledge distill model and even outperform large teacher model by leveraging information from resource-rich source domain. 
And the improvement is more obvious when the amount of data is smaller.
This shows L2A can effectively help BERT knowledge distillation for data-scarce domains.

\noindent \textbf{Effect of different source domains}.
We further proceed to examine the impact of similarity between the source and target domains. We take ``Home'', ``Electronics'' as source data respectively and take ``Watches'' as target data. To keep fair comparisons, we both choose 5000 instances for source data to help the same data-scarce target. 
We find that using ``Electronics'' instead of ``Home'' as source domain achieves better results by 0.5\% which shows transferring between similar domains leads to better performance.

\begin{table}[t!]
\caption{Ablation study on different distillation objectives.}
\centering
\begin{tabular}{l l l l l}
\hline
& $\mathcal{L}_{KD}$ & w/o $\mathcal{L}_{att}$ & 
w/o $ \mathcal{L}_{hidden}$ & w/o $\mathcal{L}_{dark}$ \\\hline
ACC & \textbf{0.7798} & 0.7563 & 0.7629 & 0.7647 \\
F1  & \textbf{0.7614} & 0.7506 & 0.7608 & 0.7433 \\
\hline
\end{tabular}
\label{tab:kd_loss}
\end{table}

\noindent \textbf{Effect of different sample size}.
We also conduct experiments to show how the number of generated augmented sentences per original sentence affects performance. We find that augmenting target domain data yields large performance boosts. By augmenting more source domain data has slightly better performance. We suspect that, if the source domain is relatively large, more source domain data does help the target but may not be very significant.

\noindent \textbf{Effect of different student models.} 
We use the first 4 layers of the teacher model as student initialization and find that L2A is also effective, as it improves 10\% in terms of ACC over the base student model on the text classification task. This shows the L2A method is generally helpful for the student model with different initialization setups.

\noindent\textbf{Effects of distillation objective.} To investigate the effects of distillation objectives, we compare the L2A method with its variants: the L2A without the attention layer (w/o $\mathcal{L}_{att}$),
the intermediate hidden layer (w/o $\mathcal{L}_{hidden}$), and dark knowledge distillation (w/o $\mathcal{L}_{dark}$). As in Table~\ref{tab:kd_loss}, we find that all the distillation objectives are useful, and the model achieves the best performance by combining all the objectives.

\noindent\textbf{Parameter sensitivity analysis}.
The temperature $\alpha$ in Eq.~\ref{eq:param-a} and $T$ in Eq.~\ref{eq:param-t} both control the search space surrounding the real data as analyzed in the method section. To investigate their impact on the performance of our model, we test with different temperature values on the text classification task. As shown in Figure~\ref{fig:temp}, we find it is better to set $T$ as a small value, which shows constrained search space is beneficial.
Meanwhile, we also find that as the temperature $\alpha$ becomes larger, the quality of augmented data gets worse.
This is because a large temperature encourages to generate more samples that are distant from the original data distribution, resulting in performance degradation.

\section{Conclusion}
We proposed a learning to augment method for BERT knowledge distillation to augment a data-scarce target domain with resource-rich source domains. 
We automatically augment the target data and dynamically refine the augmentation strategy based on the feedback from the student model. 
Extensive experiments demonstrate that the method significantly outperforms the competing baselines on various NLP tasks.

\section{Acknowledgments}
This research is supported by  National Natural Science Foundation of China (Grant No. 61773229 and 6201101015), Shenzhen Giiso Information Technology Co. Ltd., and Overseas Cooperation Research Fund of Graduate School at Shenzhen, Tsinghua University (Grant No. HW2018002).

\bibliography{aaai21}

\end{document}